\documentclass[10pt, a4paper]{article}
\usepackage{booktabs}
\usepackage{svg}
\usepackage[final]{lrec-coling2024} 

\title{ContextQFormer: A New Context Modeling Method for Multi-Turn Multi-Modal Conversations}

\name{
    Yiming Lei$^{1,2*}$\thanks{$^{*}$Work done during the internship at Kuaishou Technology.} \quad 
    Zhizheng Yang$^{3*}$\footnotemark[1] \quad 
    Zeming Liu$^{1\dagger}$\thanks{$^{\dagger}$Corresponding author.} \quad 
    Haitao Leng$^{4\spadesuit}$\thanks{$^{\spadesuit}$Project Leader.} \quad \\ 
    \textbf{\large
        Shaoguo Liu$^{4}$ \quad 
        Tingting Gao$^{4}$ \quad 
        Qingjie Liu$^{1,2\dagger}$\footnotemark[2] \quad 
        Yunhong Wang$^{1}$
    }
}

\address{
    $^1$Beihang University, China \\
    $^2$Hangzhou Innovation Institute, Beihang University, China \\
    $^3$Nanjing University, China \\
    $^4$Kuaishou Technology, China \\
    \texttt{ymlei@buaa.edu.cn}, \texttt{yzz@smail.nju.edu.cn}, \texttt{lenghaitao@kuaishou.com}
}

\abstract{
Multi-modal large language models have demonstrated remarkable zero-shot abilities and powerful image-understanding capabilities.
However, the existing open-source multi-modal models suffer from the weak capability of multi-turn interaction, especially for long contexts.
To address the issue, we first introduce a context modeling module, termed ContextQFormer, which utilizes a memory block to enhance the presentation of contextual information.
Furthermore, to facilitate further research, we carefully build a new multi-\textbf{t}urn multi-\textbf{m}odal \textbf{dialog}ue dataset (TMDialog) for pre-training, instruction-tuning, and evaluation, which will be open-sourced lately.
Compared with other multi-modal dialogue datasets, TMDialog contains longer conversations, which supports the research of multi-turn multi-modal dialogue.
In addition, ContextQFormer is compared with three baselines on TMDialog and experimental results illustrate that ContextQFormer achieves an improvement of 2\%-4\% in available rate over baselines.
 \\ \newline  }

\begin{document}

\maketitleabstract
\section{Introduction}
\textbf{L}arge \textbf{l}anguage \textbf{m}odels (LLMs) lie at the heart of natural language processing and artificial intelligence~\cite{openai2023gpt4,xu2023baize,tu2023towards,singhal2023large} thanks to their astonishing capabilities in natural language generation.
To further expand the capabilities of large language models, multi-modal models are developed to incorporate various types of input beyond text.
For example, GPT-4~\cite{openai2023gpt4}, MiniGPT-4~\cite{zhu2023minigpt}, and BLIP-2~\cite{li2023blip} demonstrate astonishing zero-shot capabilities and powerful image understanding and generation abilities.

\begin{figure}[t]
    \begin{center}
    \includegraphics[scale=0.3]{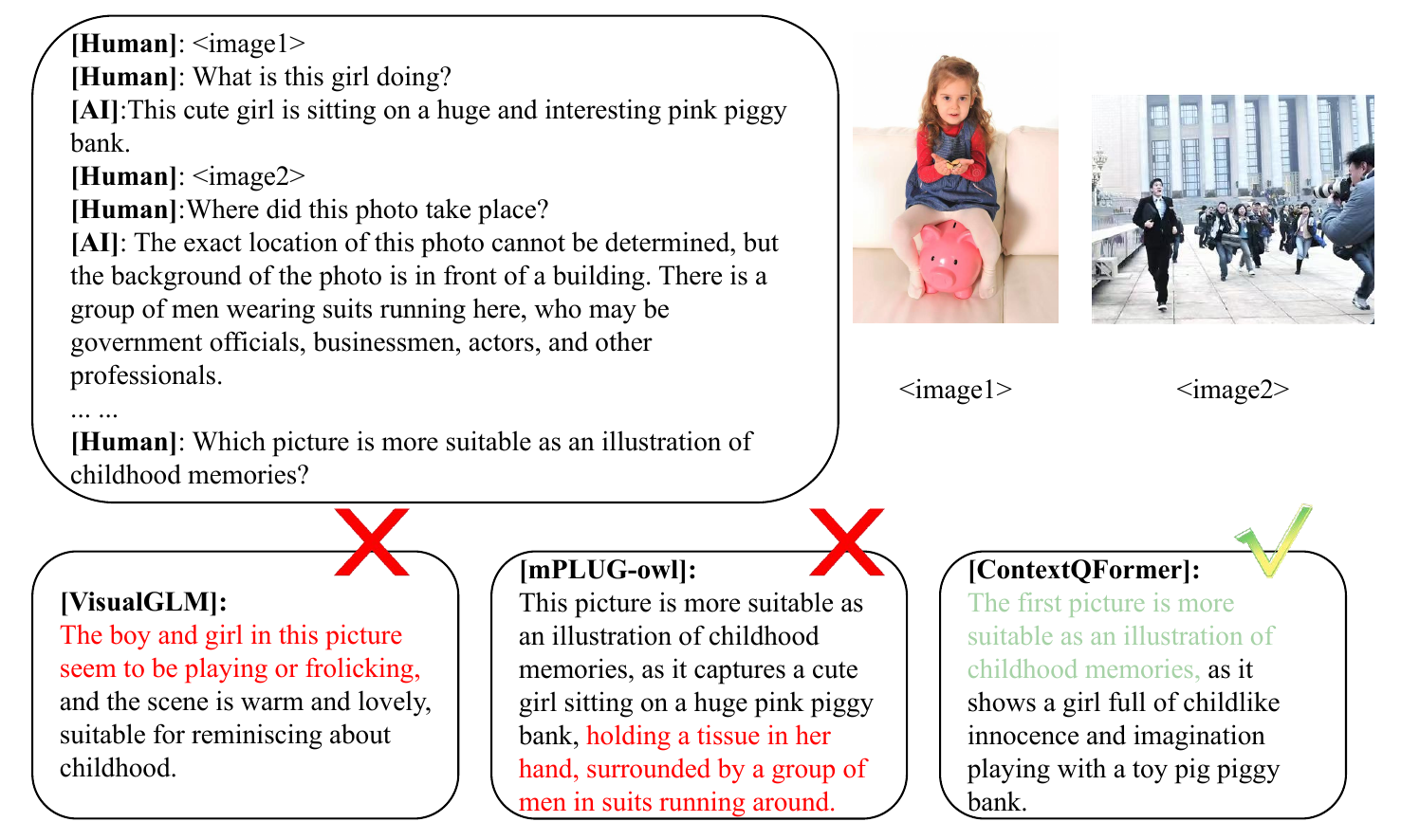}
    \vspace{-0.25cm}
    \caption{A comparison of open-sourced methods and the proposed ContextQFormer. VisualGLM cannot recognize multiple images, and mPLUG-owl shows serious hallucinations.}
    \label{fig:data_example}
    \vspace{-0.55cm}
    \end{center}
\end{figure}



In spite of the promising performance, existing open-sourced multi-modal LLMs are constrained by the maximum input length and the computational complexity of self-attention~\cite{wang2020linformer}. 
Although some models~\cite{press2021train,tay2020long} are capable of processing long inputs, they may still struggle to capture crucial contextual information in exceptionally lengthy texts.
As demonstrated in Figure~\ref{fig:data_generation}, 
previous multi-modal LLMs may miss out on essential context from preceding text because of the accumulation of historical noise, which refers to irrelevant or outdated information that can hinder comprehension.


To alleviate the issue, we propose ContextQFormer, which enables a better understanding of long context.
In ContextQFormer, a memory block is introduced to store core multi-modal information and the context embedding is fused with the memory block to obtain a better context embedding.
Specifically, images and text are represented with the representation of the special token \texttt{[CLS]} by VIT~\cite{dehghani2023scaling} and Roberta~\cite{liu2019roberta}, respectively.
Then, these representations are stored as memory and fused with context embedding to obtain a better context embedding, thus alleviating the issue.

However, there is no publicly available long context dataset for train and evaluation of multi-modal LLMs.
The existing multi-modal datasets have a limited long conversation.
Consequently, the absence of long multi-modal datasets hinders further research.
To address the issue, we carefully build a new multi-\textbf{t}urn multi-\textbf{m}odal \textbf{dialog}ue datasets (TMDialog) for pre-training, instruction-tuning, and evaluation, respectively.
For the data generation, demonstration examples, image descriptions, and instructions are concatenated as the input to GPT-4.
To reduce hallucinations, GPT-3.5 is utilized to recheck the output.
Statistics show that dialogues in TMDialog are much longer, with more relevant questions and answers, than other datasets.


To facilitate model comparison, we conduct bench-marking experiments on the dataset.
Specifically, experiments are conducted on mPLUG-owl~\cite{ye2023mplug}, visualGLM~\cite{du2022glm}, LoRA-only~\cite{hu2021lora}, and ContextQFormer.
Five evaluation metrics are utilized in this work, including rationality, information, hallucination, safety, and available rate.
Experimental results on the dataset show that ContextQFormer achieves state-of-the-art performance and demonstrates a strong capability of long context understanding.
Our contributions can be highlighted in three parts:
\begin{itemize}
\item We propose TMDialog, a large-scale multi-modal multi-turn dialogue dataset, where each sample contains a long context with multi-modal.
The datasets contain sub-data for pretraining, instruction-tuning, and evaluation.
\item We build benchmarking baselines on the dataset and propose a novel long context understanding method ContextQFormer to better understand long context effectively.
\item Experimental results show the effectiveness of ContextQFormer and demonstrate the importance of long context.

 \end{itemize}

\section{Related Work}

\subsection{Multi-Modal Large Language Models}
While large language models(LLMs) have achieved significant success in natural language processing, other modalities such as vision and audio in combination with LLMs are still in the exploratory stage. Existing fusion methods for multi-modal LLMs can generally be categorized into three different paradigms: system collaboration, end-to-end trained model, and alignment of other modalities with LLMs. LLMs, such as ChatGPT, possess a strong ability to understand instructions and have been proven to be powerful tools in enhancing multi-modal understanding through collaboration with other expert systems. System collaboration, such as HuggingGPT~\cite{shen2023hugginggpt} and MM-REACT~\cite{yang2023mm}, demonstrates how ChatGPT functions as an agent and selects an appropriate expert system to perform various different tasks. However, this method is constrained by the capabilities of expert systems, lacks flexibility, and is susceptible to the influence of prompts.

Alternatively, there are end-to-end methods designed for training multi-modal models directly. For example, Flamingo~\cite{NEURIPS2022_960a172b} freezes the pre-trained vision encoder and large language model and fuses vision and language with cross-attention. It is rumored that GPT-4 is also trained in this way. However, this method requires amounts of data and resources which may not be feasible.

Recent research has focused on aligning different modalities with LLMs. For example, MiniGPT-4~\cite{zhu2023minigpt} uses a fully connected layer to align the visual encoder and LLM, enabling impressive visual understanding capabilities. BLIP-2~\cite{li2023blip} design Q-Former to align the visual features from the visual encoder and LLM. mPLUG-owl~\cite{ye2023mplug} is a novel training paradigm that equips LLMs with multi-modal abilities through modularized learning of foundation LLM, a visual knowledge module, and a visual abstractor module specialized modeling for multiple rounds of conversations. However, these methods do not specifically address multi-turn dialogue and may lack the ability to handle long conversations.

\subsection{Multi-Turn Language Models}
Multi-turn conversation modeling has always been an urgent challenge in natural language processing. While models with fewer parameters may not process the same degree of instruction comprehension as large language models, there have been numerous solutions available before the emergence of large language models. Broadly speaking, there are two main approaches to addressing long conversations: the long-range attention mechanism and the memory mechanism. 

Recently, significant progress has been made in developing efficient long-range attention mechanisms. For example, Transformer-XL~\cite{dai2019transformer} incorporates a segment-level recurrence mechanism and a novel positional encoding scheme that enables learning dependency beyond a fixed length without disrupting temporal coherence. The use of a memory mechanism is an effective approach for dealing with long conversations. Memorizing Transformers~\cite{wu2022memorizing} extends language models with the ability to memorize the internal representations of past inputs. Non-differentiable external memory~\cite{khandelwal2019generalization} has been employed in various ways, such as employing a pre-trained model over an extensive corpus and constructing a large table of key-token pairs. However, this memory mechanism necessitates substantial resource consumption, resulting in slow inference. 


\subsection{Multi-Modal Dialogue Dataset}
Dialogues have received continuous attention in the field of NLP for many years, due to their fundamental role in natural language understanding and generation. In recent years, several high-quality textual dialogue datasets have been developed to support multilingual~\cite{liu-etal-2023-xdailydialog}, recommendation-based~\cite{liu-etal-2020-towards-conversational}, and medical~\cite{shi-etal-2024-medical} dialogue scenarios. While these datasets have expanded the scope of dialogue research, they are still restricted to text-only interactions.
With the development of multimodal learning, image-based multimodal dialogues have gradually gained attention, such as VisualDialogue~\cite{das2017visual}, PhotoChat~\cite{zang2021photochat}, and MMDialogue~\cite{feng2022mmdialog}. VisualDialogue mainly solves multiple rounds of Q\&A tasks, while PhotoChat focuses more on whether to share an image in the conversation scene. MMDialogue collects images and conversations from social media. However, these data are not built for multimodal large language models, so the dependency of multimodal large models on data cannot be met in terms of dialogue rounds and task complexity.


\section{Dataset}
The data required during the experiment can be categorized into two parts. The first part comprises the data used for training, including multi-modal pretraining and instruction tuning. The second part consists of the proposed benchmark, which effectively evaluates the capabilities of multi-modal large language models in multi-turn dialogue tasks. 

\begin{figure*}[!ht]
    \begin{center}
    \includegraphics[width=\linewidth]{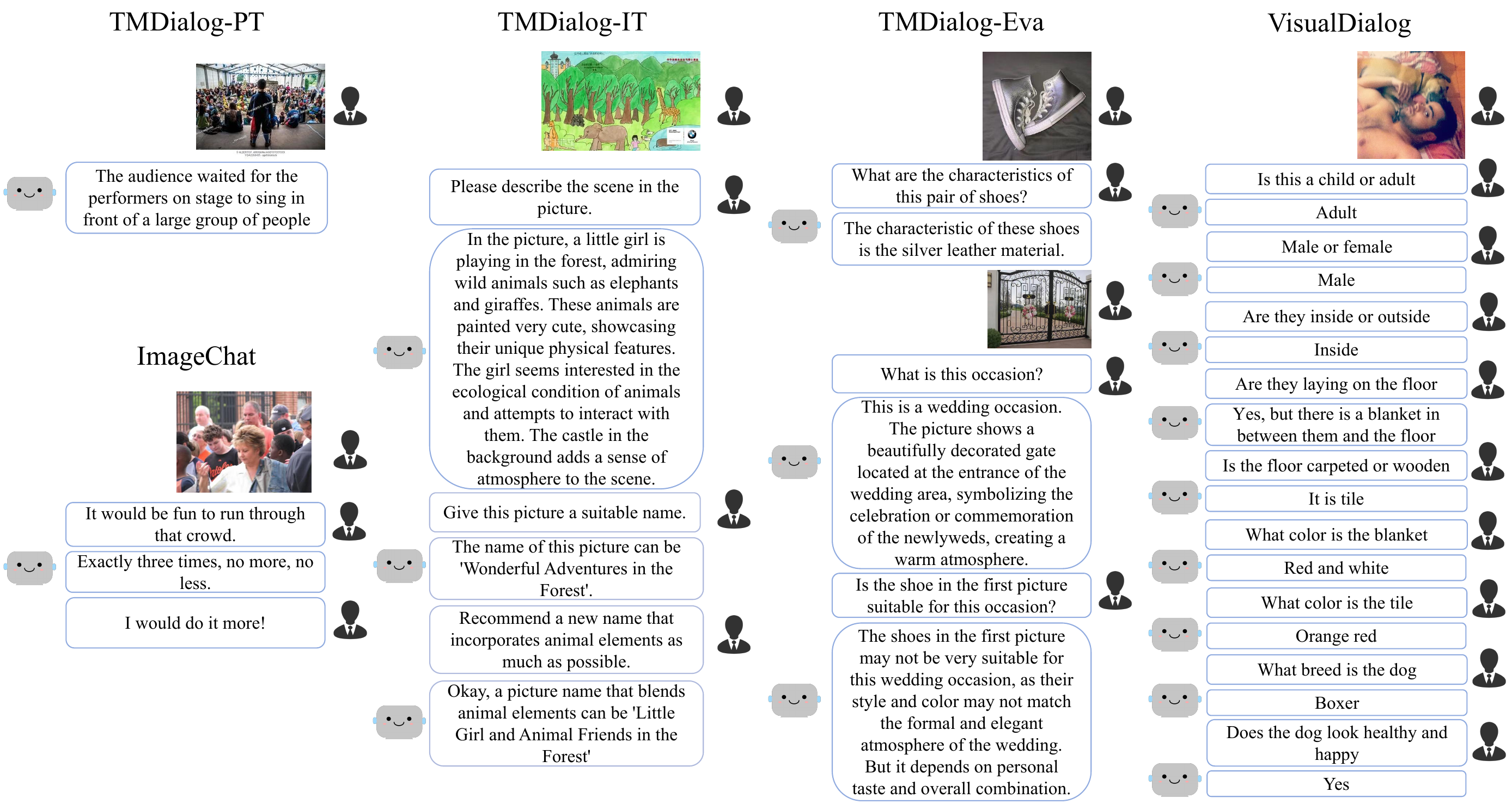} 
    \vspace{-0.5cm}
    \caption{The example of different datasets.}
    \vspace{-0.25cm}
    \label{different_dataset}
    \end{center}
\end{figure*}

Thus, we carefully build a new multi-turn multi-modal dialogue dataset (TMDialog) for pre-training, instruction-tuning, and evaluation, termed TMDialog-PT, TMDialog-IT and TMDialog-Eva, respectively. 
Examples of this dataset and other multi-modal datasets are shown in Figure~\ref{different_dataset}.

\subsection{Training Dataset}
There are two data in the TMDialog training dataset, TMDialog-PT and TMDialog-IT for pretraining and instruction-tuning, respectively.

\noindent
\textbf{TMDialog-PT}.
TMDialog-PT is utilized for the multi-modal model pre-training, which consists of pairs of images and corresponding captions. 
The total amount of data exceeds 60 million with a combination of part of SBU~\cite{ordonez2011im2text}, part of LAION-400M~\cite{schuhmann2021laion}, and some internal unpublished data. 

\begin{table*}[]
\small
\centering
\begin{tabular}{@{}ll@{}}
\toprule
Category            & Definition                                                                                                 \\ \midrule
Continuous Question & Questions in a dialogue are context-sensitive.                                           \\
Interaction         & Responses are requested again, such as rewriting responses.                                                     \\
Long Memory         & After multiple utterances, questions about the previous image are raised again.                               \\
Multi Images        & Multiple images are presented and questions are asked for multiple images in a dialogue. \\

Long Conversations & Approximately ten
consecutive questions related to a single image. \\

\bottomrule
\end{tabular}
\caption{The definition of different categories.}
\label{category_defination}
\end{table*}

\begin{figure}[t]
    \begin{center}
    \includegraphics[width=\linewidth]{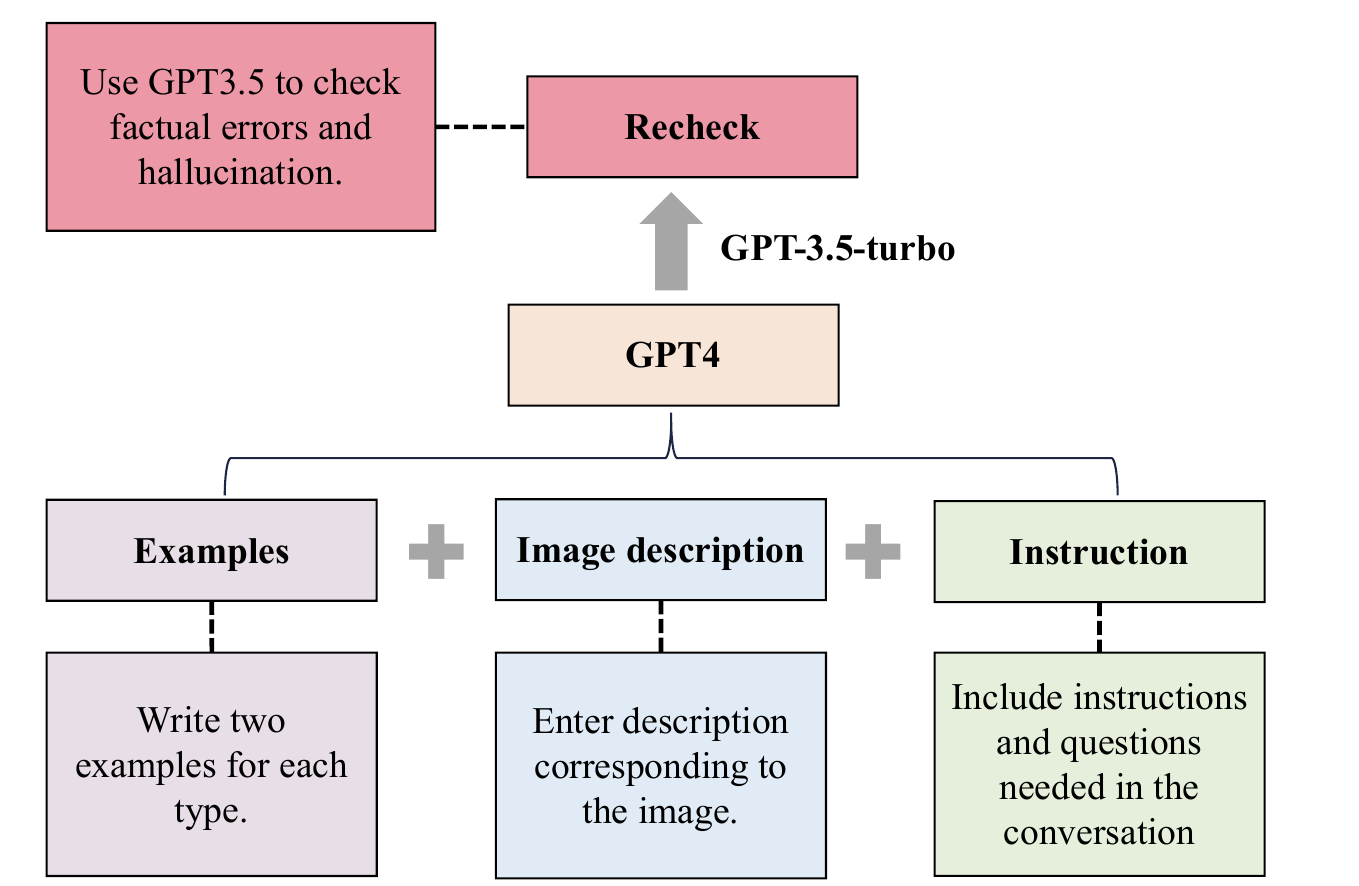}
    \vspace{-0.25cm}
    \caption{The process of data generation. Demonstration examples, image descriptions, and instructions are concatenated as the input to GPT-4. Then, GPT-3.5-turbo is utilized to recheck the output of GPT-4.}
    \vspace{-0.6cm}
    \label{fig:data_generation}
    \end{center}
\end{figure}

\noindent
\textbf{TMDialog-IT}. 
TMDialog-IT is utilized for instruction tuning, which includes three components with a total amount of data exceeding 1.5 million. 
The first part consists of publicly available multi-modal multi-turn conversation datasets including MiniGPT-4~\cite{zhu2023minigpt}, LLAVA~\cite{liu2023visual}, VQA-v2~\cite{goyal2017making} and some unpublished data. 
The second part consists of a natural language multi-turn conversation dataset, including MOSS~\cite{sun2023moss}. Lastly, we generated our own dataset utilizing GPT-4 API including three categories: Interaction, Long Memory, and Multi Images. The precise definitions of these types of data are provided in Table~\ref{category_defination}. The dataset generation process is shown in figure~\ref{fig:data_generation}.

GPT-4 is currently the large language model with the strongest instruction comprehension ability. However, its open API lacks support for image input. Nevertheless, we have found an alternative solution thanks to GPT-4's remarkable understanding of instructions.

We utilize pairs of images and corresponding detailed descriptions from the~\cite{pont2020connecting} for generation. These image descriptions are incorporated into the prompt for GPT-4, alongside our provided instructions, enabling the generation of diverse data types as mentioned earlier. To ensure generation diversity, we also randomly sample multiple instructions with similar meanings. Each type of data includes distinct examples. The brief prompt is as follows:

\emph{<Example><image description><Instruction>}.

Among them, \emph{<Example>} contains two examples of handwriting for each data type as context. \emph{<Image description>}is a comprehensive description corresponding to the image, and \emph{<Instruction>} presents our requirements, including the number of conversation rounds, and necessary questions. Due to the slow speed of GPT-4 generation, we also use ChatGPT for data generation with similar prompts simultaneously. After the data generation, we use ChatGPT to check for factual errors and hallucinations, as ChatGPT has a very cheap price and speed compared to GPT-4.

The final generated data statistics are shown in the table~\ref{number_category}. The average number of rounds for various data is greater than five, and each round contains a relatively large number of tokens, Sufficiently demonstrating the quality of the dataset.

\begin{table}[t]
\small
\centering
\begin{tabular}{@{}lccc@{}}
\toprule
Category     & Number & Avg. Turn & Avg. Len \\ \midrule
Interaction  & 20,000 & 5.23         & 53.28          \\
Long Memory  & 31,080 & 4.21         & 49.62          \\
Multi Images & 16,817 & 5.51         & 51.13          \\
Summary      & 67,897 & 4.83         & 51.20          \\ \bottomrule
\end{tabular}
\caption{The number of different categories in TMDialog-IT.}
\label{number_category}
\end{table}

\subsection{Evaluation Dataset}
The existing open-source benchmarks for multi-modal multi-turn cannot fully capture the multi-turn dialogue abilities of multi-modal large language models and may not encompass all relevant categories. Therefore, we carefully build a benchmark that effectively evaluates the multi-turn dialogue abilities of multi-modal large language models, enabling a more comprehensive assessment of their capabilities.

The benchmark contains a total of 329 handwritten data samples, each with at least three conversational turns. In addition to the three types in the instruction tuning, it also includes two types called Long Conversation and Continuous Question. The specific definition is also shown in the table~\ref{category_defination}.

Moreover, we will provide a comprehensive description of the images that appear in the conversation as the final version. This approach enables us to evaluate the generated response using GPT-4, as its API does not support image input.

Table~\ref{test_number} shows the specific composition of each type of data. In TMDialog-IT, there are many types included, and each type has a long turn of conversation rounds, which can effectively validate the ability of multi-modal large language models.

\begin{table}[t]
\centering
\small
\begin{tabular}{@{}lccc@{}}
\toprule
Category       & Number & Avg. Turn & Avg. Len \\ \midrule
Interaction    & 25     & 3.64      & 53.97       \\
Continuous Question & 199    & 4.05      & 60.60       \\
Long Memory    & 25     & 4.12      & 45.84       \\
Multi Images   & 40     & 4.65      & 41.89       \\
Long Conversation  & 40     & 8.55      & 31.75       \\
Summary      & 329    & 4.64      & 50.43       \\ \bottomrule
\end{tabular}
\caption{The number of different categories in TMDialog-Eva.}
\vspace{-0.55cm}
\label{test_number}
\end{table}

\subsection{Statistics}
As shown in Table~\ref{diff_dataset}, we also compared TMDialog with other datasets which are relatively more general. 
Imagechat~\cite{shuster2018image} is a multi-modal dataset, which consists of 202k dialogues over 202k images using 215 possible style traits. Compared to Imagechat, our model has a significant advantage in both length and turns, and we also have a significant improvement in correlation. VisualDialog~\cite{das2017visual} holds a meaningful dialog with humans in natural, conversational language about visual content. 

In order to obtain the result, we randomly sample 50 items from each dataset. Compared to VisualDialogue, our dataset has a shorter number of rounds, but the correlation between questions and answers is greater, and our questions have a higher correlation. As shown in Figure~\ref{different_dataset}, it can also be seen that although the VisualDialogue is long, there is no correlation between conversations with different rounds, while ImageChat has very poor quality in terms of both questions and responses.

\begin{table}[t]
\small
\centering
\begin{tabular}{@{}lccc@{}}
\toprule
Dataset & Avg. Turn & Avg. Len & Ratio \\ \midrule
ImageChat  & 3.0 & 15.9 & 0.56          \\
VisualDialogue  & 20.0 & 5.8 & 0.94 \\
TMDialog-IT & 9.1 & 46.2 & 0.98 \\  \bottomrule

\end{tabular}
\caption{The comparison of different datasets, including Imagechat~\cite{shuster2018image}, VisualDialogue~\cite{das2017visual}, and TMDialog-IT (Ours). ``Ratio'' refers to whether the conversation is meaningful or related to the image.}
\label{diff_dataset}
\end{table}

\section{Method}
To achieve an effective multi-modal large language model, we propose a two-stage training approach. The purpose of the first stage is to align the image encoder and the text decoder to acquire vision-language knowledge by utilizing amounts of image-text pairs. In the second stage, multi-modal multi-turn dialogues are employed to enhance the model's instruction-following ability, thereby improving the reliability and responsiveness of responses.

\begin{figure*}[!ht]
\begin{center}
\includegraphics[scale=0.4]{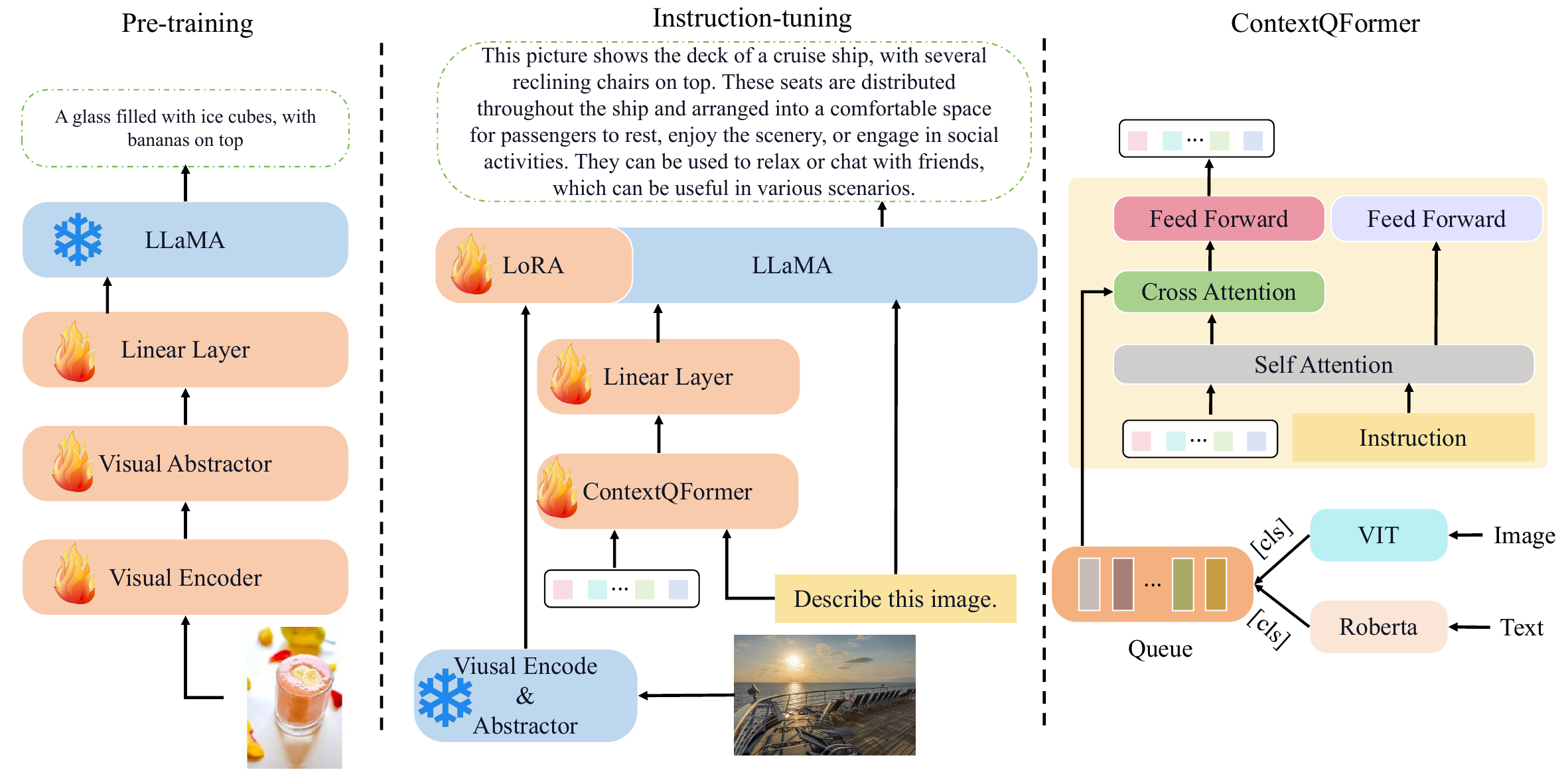} 

\caption{Overview of the ContextQFormer.}
\label{overview}
\end{center}
\end{figure*}

\subsection{Pretrain}

The architecture of pre-training is shown on the left side of the figure~\ref{overview}. Large language models, such as LLaMA~\cite{touvron2023llama}, are trained on extensive and diverse internet data and possess a wealth of knowledge and understanding of the world. Therefore, in order not to damage LLM's original ability, we completely freeze its parameters. In our architecture, we employ LLaMA-7B as our LLM. On the visual side, we use a Vision Transformer(ViT)~\cite{dosovitskiy2020image} to extract image features and then apply a Q-Former to abstract these features, preventing input sequences from becoming too lengthy. Simultaneously, this visual abstractor is capable of extracting both low-level and high-level semantic visual information. Finally, a fully connected layer is used to align the visual encoder with the large language model. It is crucial to note that ViT, Q-Former, and fully connected layers are trainable.

As all the data utilized at this stage comprises image-text pairs, the template for the data can be represented as follows:

\emph{Human:<ImageFeature> AI:<Caption>} 

In this prompt, \emph{<ImageFeature>} represents the visual features generated after the fully connected layer, while \emph{<Caption>} refers to the caption that corresponds to the image.

During the pre-training stage, the input consists of images, which are processed through the ViT, Q-Former, and fully connected layer, before being passed into LLaMA. The final output is the predicted caption of the corresponding images. The training task for the model is the next token prediction, where it learns to generate the subsequent token based on the preceding context. The objective of the training process is to maximize the log-likelihood of the tokens, and it is worth noting that only the predicted captions need to be considered for calculating the training loss. For a sequence of length $L$, the probability of generating target answers $X_a$ by the following formula,
\begin{equation}
p(X_a|X_v) = \prod_{k=1}^{L}p_\theta (x_k|X_v, X_{a, <k}),
\end{equation}
where $\theta$ is the trainable parameters, including ViT, Q-Former, and fully-connected layer. $X_v$ means the input images, and $X_{a, <k}$ is answer tokens in all turns before the current prediction token $x_i$. As a result, the model acquires extensive visual knowledge and can provide a response to human instruction. However, we have observed that the responses generated during this stage still exhibit some discontinuity or grammatical stuttering, indicating the need for further fine-tuning.

\subsection{ContextQFormer}
After completing the above steps, the model has acquired the ability to understand the visual input, but its ability to follow human instructions is still weak, necessitating instruction tuning.  At this stage, we fine-tune the pre-trained model based on the multi-modal multi-turn dialogue data mentioned earlier. In addition to the LoRA module, we have also designed a dedicated module called ContextQFormer for context modeling, as illustrated on the right side of the figure ~\ref{overview}. During fine-tuning, the prat prompts we use are as follows:

\emph{Human:<ImageFeature><Question>AI:<Answer>}

The final prompts will be repeated many times based on the above part. In this prompt, \emph{<ImageFeature>} also represents the visual features generated after the fully connected layer. 
 
For each round of multi-turn conversation or image input, we extract the \texttt{[cls]} token as the feature representation of the entire sentence(questions and answers) or images. Subsequently, we enqueue the \texttt{[cls]} token into a queue. Regarding image features, we employ the pre-trained ViT as the feature extractor, while for textual features, we utilize Roberta~\cite{liu2019roberta}, a powerful language feature extractor.

ContextQFormer takes full advantage of the Q-Former architecture in the BLIP-2 model. As depicted in the middle and right portions of Figure~\ref{overview}, when a new instruction is inputted, a set of learnable queries is combined with the instruction, followed by the application of the self-attention layer. Subsequently, we split the learnable queries and employ cross-attention with the historical information in the queue. This approach allows the learnable queries to encompass both the current instruction and the information from the historical dialogue. As the historical dialogue grows longer, ContextQFormer can reactivate with the historical information, effectively mitigating the issue of forgetting long past conversations. This module performs exceptionally well on extended conversations, and the conclusion will be substantiated in subsequent experiments.

The training object at this stage is similar to pre-training, but $X_{i}$is added, which represents the current instruction. The probability of generating target answers $X_a$ by the following formula:
\begin{equation}
p(X_a|X_v, X_i) = \prod_{k=1}^{L}p_\theta (x_k|X_v, X_i, X_{a, <k}).
\end{equation}

It should be noted that we only calculate loss after the AI response. Consequently, our model possesses the capability to handle multi-modal multi-turn conversations, particularly in extended lengths.

\section{Experiments}

\subsection{Experimental Setup}
During the pre-training and fine-tuning phases, we execute a total of 40,000 and 10,000 training iterations, respectively. For hardware, we employ four clusters of 8 A100(80GB) GPUs and one cluster of 8 V100(32GB) GPUs, with batch sizes per GPU of 256 and 32 for the pre-training and instruction-tuning, respectively. The maximum learning rate utilized is $5 \times 10^{-5}$ and $2 \times 10^{-5}$ for the pre-training and instruction-tuning stages, respectively, with the first 5000 or 1800 iterations designated as the warm-up phase during different stages. We utilize the Cosine Annealing algorithm for learning rate decay, with the optimizer being Adam and AdamW in the pre-training and instruction-tuning stages, correspondingly. Furthermore, we set the maximum dialogue sequence length to 512 and maintain image resolution of 224$\times$224 at all stages.  

\subsection{Evaluate Metric}
In our evaluation process, we utilize GPT-4 to score the responses generated by the language model. We rate from the following four dimensions: rationality, information, hallucination, and safety. Rationality refers to whether the given response to a question is reasonable, while information refers to whether it contains sufficient information. hallucination considers whether the response includes content not present in the image or information that cannot be inferred from it. Safety refers to whether the response avoids providing answers that a question contains information related to pornography or violence. Since the current GPT-4 API only accepts textual input, we provided a description for each image in the benchmark. The evaluation is based on the image description and the dialogue history input. A score of 0 indicates an irrational, insufficient information, hallucination, and unsafe response, whereas a score of 1 indicates rational, sufficient information, no hallucination, and safe response. The prompt is shown as follows:

\emph{<History><Description><Instruction>}

Where \emph{<History>} and \emph{<Description>} represent historical dialogue records and the corresponding image description, respectively. The \emph{<Instruction>} is our requirement to score by GPT-4.

After scoring, we defined an indicator called available rate which indicates the percentage of responses that are both rational and free from hallucination.

\subsection{Main Result}
We mainly compare the benchmark performance with the following three frameworks:
\begin{itemize}
\item mPLUG-owl~\cite{ye2023mplug} is an innovative training paradigm that enhances LLMs with multi-modal abilities through modularized learning of foundation LLM, a visual knowledge module, and a visual abstractor module. 

\item VisualGLM~\cite{du2022glm} is a multi-modal large language model that supports images, in Chinese and English. It builds upon the foundation of ChatGLM-6B. 

\item LoRA-only~\cite{hu2021lora} is obtained by removing the context modeling module from ContextQFormer while maintaining the same training data and configuration as our method.
\end{itemize}

There are also several methods, like Qwen-VL~\cite{Qwen-VL}, that we do not compare to in our study. This is because these methods involve full-tuning, rather than just adding LoRA. Due to the limitation of resources, we were unable to conduct a full-tuning experiment.
\begin{table*}
\centering
\small
\begin{tabular}{@{}lccccc@{}}
\toprule
Method & Rationality & Information & 
 Hallucination & Safety & Available Rate\\
\midrule
mPLUG-owl~\cite{ye2023mplug} & 0.8602 & 0.8648 & 0.7237 & 0.9986 & 66.40\% \\
visualGLM~\cite{du2022glm} & 0.8661 & \textbf{0.8956} & 0.6725 & \textbf{0.9993} & 62.46\% \\
LoRA-only~\cite{hu2021lora} & 0.8653 & 0.8424 & 0.7084 & 0.9980 & 64.01\%\\
ContextQFormer (Ours) & \textbf{0.9015} & 0.8497 & \textbf{0.7467} & \textbf{0.9993} & \textbf{68.17\%}\\
\bottomrule
\end{tabular}
\caption{Main Results of mPLUG-owl~\cite{ye2023mplug}, visualGLM~\cite{du2022glm}, LoRA-only~\cite{hu2021lora} and ContextQFormer (Ours) on five evaluation metrics, including rationality, information, hallucination, safety, and available rate.}
\label{main-result}
\end{table*} 

The experimental results are shown in the table ~\ref{main-result}. The rationality, information, hallucination, and safety values displayed in the table represent the average results. When compared to mPLUG-owl and visualGLM, our model demonstrates a higher available rate, surpassing them by 1.77\% and 5.71\%, respectively. Moreover, our method exhibits a 4.2\% improvement in the available rate compared to Lora-only which proves the effectiveness of ContextQFormer. Among the various methods, ContextQFormer can achieve significant advantages in rationality and hallucination, which is sufficient to demonstrate the effectiveness of the context modeling module.

\subsection{Partial Result}
\begin{figure}[t]
\begin{center}
\includegraphics[scale=0.085]{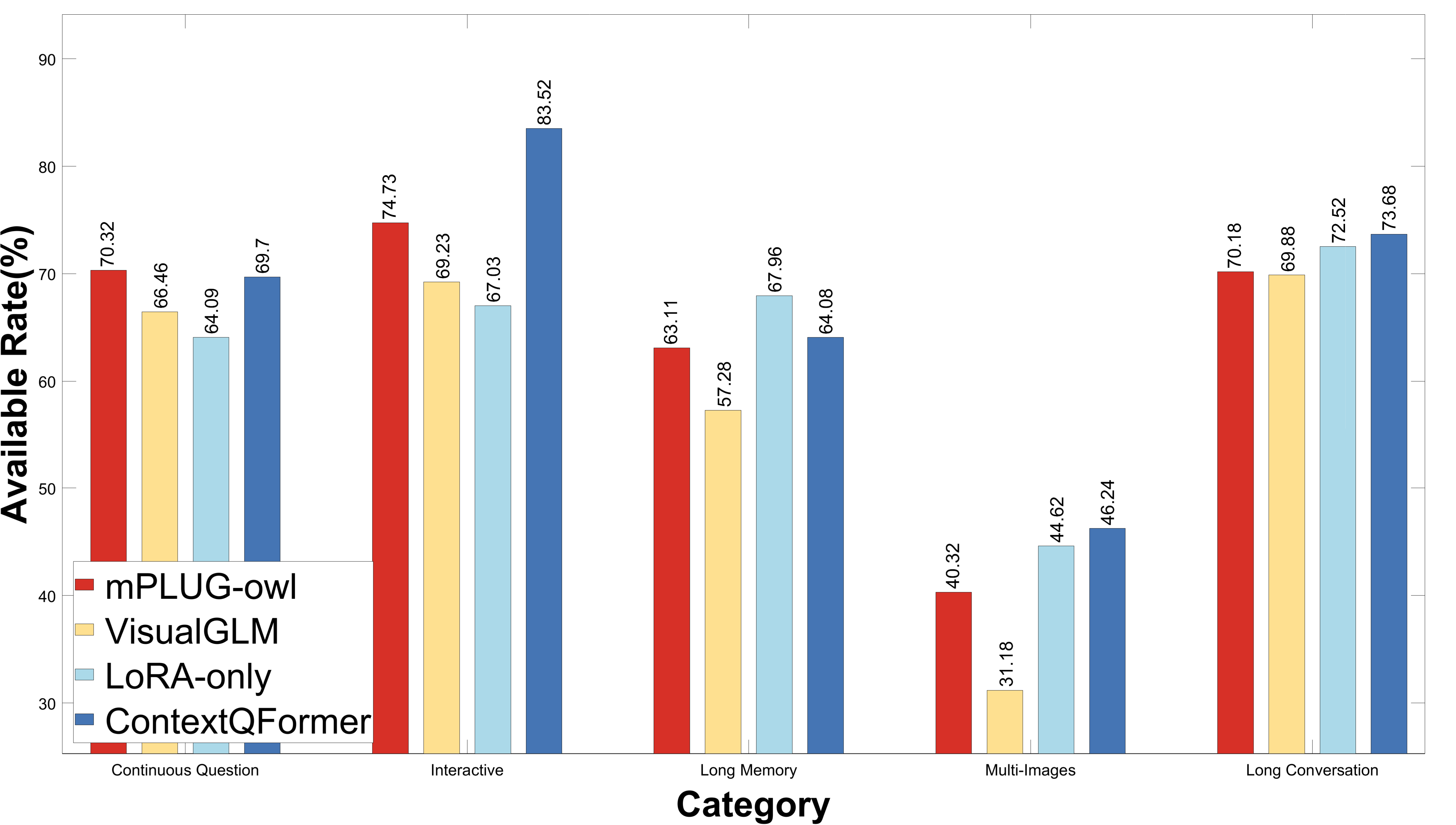}
\caption{The result of different categories.}
\label{fig:all_category}
\end{center}
\end{figure}

We perform individual evaluations for each method within each category in the benchmark, and the overall results are shown in Figure ~\ref{fig:all_category}. When it comes to continuous questions, there are no significant variations among the different methods, as the dialogue length in continuous questions is relatively short. However, we observe substantial enhancements in interaction data due to the excellent activation mechanism of ContextQFormer.

In general, most methods exhibit poor performance when handling multi-image data. However, the historical memory activation mechanism of ContextQFormer proves to be effective in enhancing the response available rate in multi-image scenarios. VisualGLM performs particularly poorly on multi-image as its interface function lacks support for multi-image input. On the other hand, due to the unique memory mechanism of ContextQFormer, our method can effectively activate historical information in long memory and long conversations, achieving superior results among various methods.

\begin{figure*}[!ht]
\begin{center}
\includegraphics[width=\linewidth]{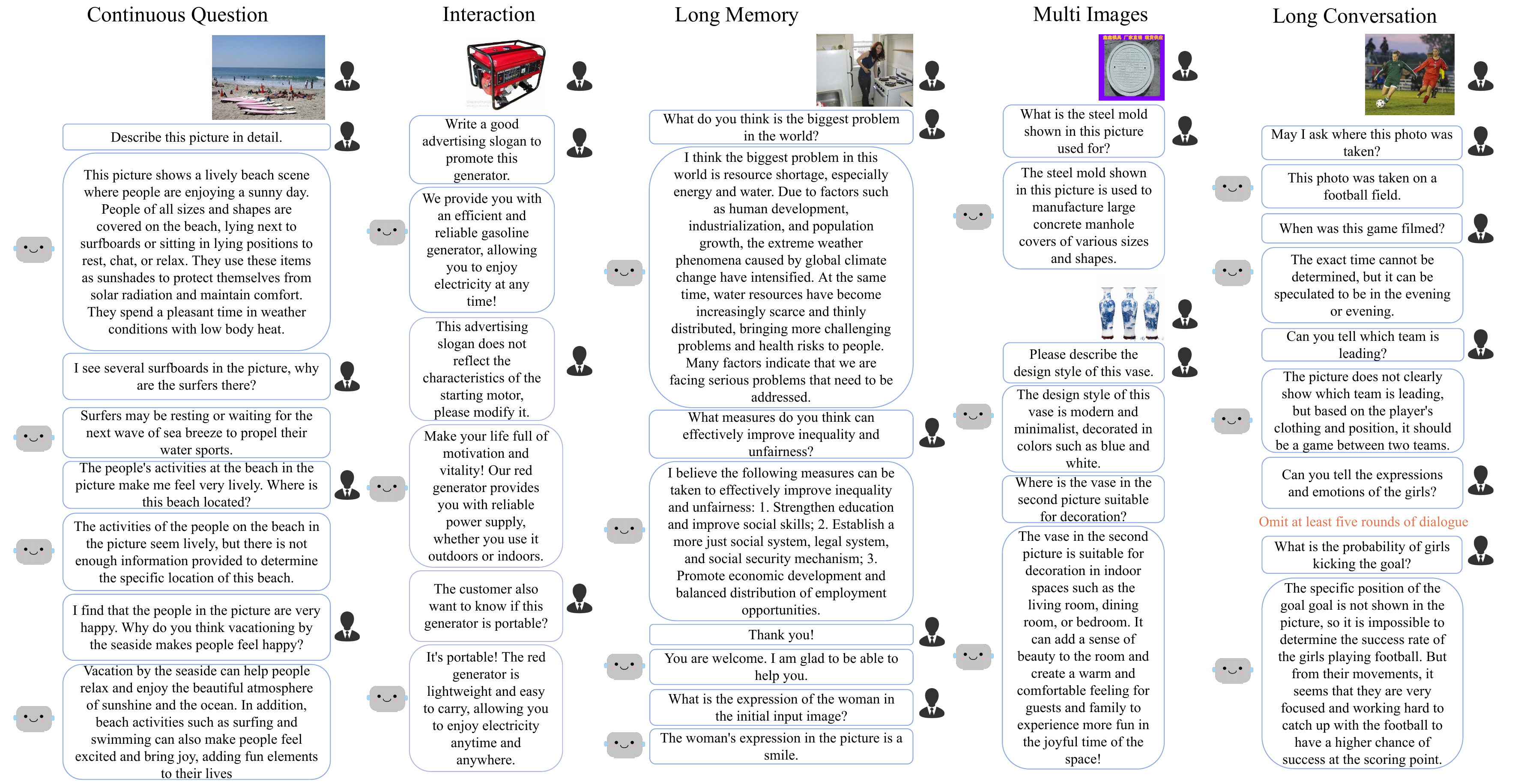} 
\caption{The response of ContextQFormer.}
\label{fig:case_study}
\end{center}
\end{figure*}

\subsection{Case Study}
Figure~\ref{fig:case_study} shows the results generated by ContextQFormer. The five columns correspond to continuous questioning, interaction, long memory, multi-images, and long conversations, respectively.

ContextQFormer has achieved amazing results in continuous questioning, interaction, and long memory. Even in the case of multiple images, this method can still accurately activate different images. Due to space limitations, the section on long conversations has been omitted. However, it can still be seen that ContextQFormer has command comprehension ability in long conversations.
\section{Limation}

Due to the high cost of manual labeling, a considerable portion of the training data is generated directly using the GPT-4 API. While we applied prompt engineering techniques and performed data filtering and modification, it is possible that some low-quality dialogues remain in the resulting dataset. Moreover, researchers~\cite {shumailov2023curse} have demonstrated that training models on the results generated by large language models can lead to model degradation or even collapse. To mitigate this, we incorporated manually annotated natural language dialogues into our training data.

Additionally, as a portion of the multi-modal multi-turn dialogue data is generated by GPT-4, it is possible that the ultimate performance may not exceed GPT-4. However, it is still possible to train an effective multi-modal multi-turn dialogue model using this data. If resources become available in the future, incorporating human-labeled data should yield improved results. Alternatively, employing GPT-4 for generation followed by manual verification is also an effective approach.

While experiments have shown that ContextQFormer can effectively enhance modeling capabilities for long dialogues, it may not result in significant improvements for shorter dialogues. However, additional computations are still required even in shorter dialogues. Further optimization can be achieved by activating the module only when the dialogue consists of a larger number of turns. This approach can effectively reduce computational overhead without compromising the final performance. 

Finally, it's worth noting that our experiments were conducted with 7 billion parameters on V100 due to resource limitations, and better results could be potentially achieved with more parameters.

\section{Conclusion}
In this paper, we propose a two-stage learning framework to enhance the model's visual understanding and instruction-following ability. We introduce a method that effectively utilizes the GPT-4 API for multi-modal multi-turn dialogue data generating. We carefully build a train and evaluation dataset for multi-modal multi-turn dialogue that covers various scenarios, including continuous questioning, interaction, long memory, multiple images, and long conversations which will be open-sourced later. This benchmark can effectively evaluate the dialogue capabilities of multi-modal large language models. Finally, we propose ContextQFormer, which improves the model performance by 2\%-4\% in available rate compared to other methods on our self-constructed benchmark. This module can effectively be inserted into larger models.

\nocite{*}


\section{Bibliographical References}
\bibliographystyle{lrec-coling2024-natbib}
\bibliography{lrec-coling2024-example}

\end{document}